RESEARCH ARTICLE

# Exploiting All Samples in Low-Resource Sentence Classification: Early Stopping and Initialization Parameters

**HONGSEOK CHOI[1] AND HYUNJU LEE[1,2], (Member, IEEE)**
[1]School of Electrical Engineering and Computer Science, Gwangju Institute of Science and Technology (GIST), Gwangju 61005, South Korea
[2]Artificial Intelligence Graduate School, Gwangju Institute of Science and Technology (GIST), Gwangju 61005, South Korea

Corresponding author: Hyunju Lee (hyunjulee@gist.ac.kr)

This work was supported in part by the National Research Foundation of Korea (NRF) Grant funded by the Korean Government (MSIT) under Grant 2021R1A2C2006268; and in part by the Institute of Information and Communications Technology Planning and Evaluation (IITP) Grant funded by the Korea Government (MSIT; Artificial Intelligence Graduate School Program, GIST) under Grant 2019-0-01842.

**ABSTRACT** To improve deep-learning performance in low-resource settings, many researchers have redesigned model architectures or applied additional data (e.g., external resources, unlabeled samples). However, there have been relatively few discussions on how to make good use of small amounts of labeled samples, although it is potentially beneficial and should be done before applying additional data or redesigning models. In this study, we assume a low-resource setting in which only a few labeled samples (i.e., 30–100 per class) are available, and we discuss how to exploit them without additional data or model redesigns. We explore possible approaches in the following three aspects: training-validation splitting, early stopping, and weight initialization. Extensive experiments are conducted on six public sentence classification datasets. Performance on various evaluation metrics (e.g., accuracy, loss, and calibration error) significantly varied depending on the approaches that were combined in the three aspects. Based on the results, we propose an integrated method, which is to initialize the model with a weight averaging method and use a non-validation stop method to train all samples. This simple integrated method consistently outperforms the competitive methods; e.g., the average accuracy of six datasets of this method was 1.8% higher than those of conventional validation-based methods. In addition, the integrated method further improves the performance when adapted to several state-of-the-art models that use additional data or redesign the network architecture (e.g., self-training and enhanced structural models). Our results highlight the importance of the training strategy and suggest that the integrated method can be the first step in the low-resource setting. This study provides empirical knowledge that will be helpful when dealing with low-resource data in future efforts. Our code is publicly available at https://github.com/DMCB-GIST/exploit_all_samples.

## I. INTRODUCTION

Data scarcity is a common and long-standing issue in artificial-intelligence research [1], [2], [3]. There are several reasons for data scarcity. First, target data are difficult to collect. For example, data collection can be difficult when it contains personal information [4], [5], such as social-media profile, or has little presence on the Web, such as African languages [6], [7]. Second, data labeling is expensive. For example, labeling of medical texts requires a high level of expertise [8], [9]. Third, data may be scarce when a customized dataset needs to be created for a specific task. For example, when developing a robot to assist human workers, developers should create new data suitable for the working environment [10], [11]. Therefore, the data scarcity problem is an important topic to be addressed.

Recent deep learning techniques have demonstrated a high capacity for encoding large and complex datasets [12]. However, in low-data regimes, such as those described above, the limited data restrict the full use of deep learning [13], [14], [15]. By focusing more on local optima, the high capacity sometimes leads to overfitting because small data may not be sufficiently representative [16]. To address this problem, many studies have introduced additional data, or redesigned model architectures. For example, transfer-learning, meta- or multitask learning (MTL), semi-supervised learning (SSL), and data augmentation methods use external datasets, sets of various tasks, unlabeled samples, and knowledge resources, respectively [17], [18], [19], [20], [21]. These methods may also require model redesign in terms of network architecture or loss function [19], [21]. They have been extensively studied in conjunction with low-resource tasks. However, few studies have examined how to make good use of only a small number of labeled samples without depending on additional data or redesign techniques. This knowledge would be useful as an initial step in a low-resource setting. For example, it might improve the accuracy of the initial model in an iterative self-training process.

In this study, we constrain our low-resource setting as follows: only a small amount of labeled data is available, and model redesign techniques are not allowable. Under these constraints, we explore several possible approaches to improving model performance based on the following three aspects: i) training-validation splitting, ii) early stopping, and iii) weight initialization.

First, we explore the training-validation split. Validation-based early stopping is a common method of preventing overfitting. However, when splitting data into training and validation sets, the limited data in a low-resource setting raise an issue. When splitting data into a small training set and a large validation set, weak learning performance tends to occur, although doing so can provide a good stopping point. Conversely, when using a large training set and a small validation set, better learning performance may occur, but an ambiguous stopping point may result. Unfortunately, there is no clear consensus on this issue. In some studies, the lack of data leading to overfitting has encouraged the use of large validation sets [22]. Makarov and Clematide [23] used 100 samples for training but 1,000 samples for validation. Upadhyay et al. [24] used 500 and 1,000 samples, respectively. In contrast, Kann et al. [22] advised against the use of large validation sets in low-resource settings. In this study, we reexamine the efficacy of the validation set in a low-resource setting. We also explore how to effectively use the validation set for performance improvement.

Second, we explore early stopping, which is even more important in low-resource settings because prediction performance fluctuates during training. In addition to validation-based early stopping, we explore four non-validation stopping methods that enable the training of all available samples [25], [26]. In low-resource settings, training even a few more samples significantly impacts performance. If an appropriate stopping point was to be given, training all available samples would be more effective. We analyze the performance curve during training with various evaluation metrics: accuracy, loss, and calibration error [27]. In our empirical results, we find that the performance of the stopping method varies significantly based on the evaluation metric.

Third, we explore weight initialization. Existing low-resource approaches often require additional data. In this study, we explore easily implementable approaches that find good initial parameters with only a few labeled samples. The randomness of parameter initialization has a significant effect on a small dataset, even when using the same initialization method. We explore three approaches: two to find better initialization based on the observed validation performance and another to find better one with gradient descent [28].

In summary, we comparatively study various easily implementable approaches requiring only a few labeled samples without model manipulation. The aim is to investigate how different performance can be depending on the training strategy. (i.e., training-validation splitting, early stopping, or weight initialization). We conduct extensive experiments using various combinations of the training strategies. Based on the results of these experiments, we determine the best performing integrated method among the explored methods, which is to initialize the model with a weight averaging method and then train all samples using a non-validation stopping method. Our weight averaging method is to average multiple models that are trained on different subsets of samples. In addition, we demonstrate how this integrated method can serve as an initial step in low-resource settings. To this end, we adapt the integrated method to several other state-of-the-art (SOTA) models and observe additional performance improvement.

Section II covers related works, and the detailed methods are presented in Section III. We describe the experimental setting and the six public sentence classification datasets in Section IV. To implement a low-resource setting, approximately 30 to 100 labeled samples per class are used for each dataset. Our empirical results are presented in Section V, demonstrating that model performance significantly varies based on the training strategy even when using the same data and architecture. Section V also includes the results of applying our integrated approach to several SOTA models and demonstrates the usefulness of our approach in low-resource settings. Finally, Sections VI and VII present our discussion and conclusions. This study can be a basic step before applying additional data or modeling techniques, highlighting the importance of the training strategy in low-resource settings.

## II. RELATED WORK

A key challenge in low-resource settings is generalization. To this end, many studies have relied on the use of additional data. Pretraining on a large dataset and fine-tuning on a small target dataset are popular approaches [29]. Transfer-learning exploits pretrained parameters for the target task [17], [18], and MTL trains multiple tasks simultaneously while covering all the classes present in the test sets [19], [20]. In contrast, meta-learning aims to predict unseen classes with only a few samples after training a set of tasks [21]. SSL exploits both few labeled data and large unlabeled data [30]. Data augmentation in natural language processing (NLP) is generally based on knowledge resources (e.g., WordNet for token replacement and the Google Translate API for back-translation) [31], [32], [33]. These methods provide good, generalizable parameters using additional techniques and data. However, in this study, we explore several methods of finding better initial parameters based on only a small number of labeled samples.

Weight averaging [34], [35], [36] and mixup [37], [38] are other methods that can be applied without additional data and without architecture redesign. Weight averaging is the parameter-wise averaging of multiple fine-tuned models. Therefore, weight averaging requires more training time and does not increase the testing time. Wortsman et al. [34] studied several weight averaging methods. Their experiments demonstrated that simply averaging in a greedy manner can bring significant performance improvement; this method is named the 'greedy soup'. We compare the greedy soup with our approach. Unlike common weight averaging methods [34], [35], [36], our integrated method averages multiple models that are trained on different subsets of samples. Furthermore, we use the averaged parameters as the initialization in the phase of training all samples. The mixup method was first proposed in computer vision [37]. Mixup imposes weights to two image samples, which are randomly sampled from the training set, and trains the pixel-wise mixed image with the weight-averaged label of the two images [37]. Therefore, mixup only requires a minor modification of the loss function without requiring additional data. Sun et al. [38] proposed a mixup method for NLP. In [37] and [38], mixup boosted the model performance. We compare the mixup method in [38] with our approach. Unlike [38], our study focuses on low-resource settings and compares various methods.

A commonly used ratio range for the training-validation split is 2:1–4:1. It often jumps to 9:1 when sufficient data are present. Several studies have investigated the optimal split ratio for training and validation [39], [40], and it is generally known that the proportion of the validation set tends to be smaller when more data are available [39], [40]. However, there is no consensus about an optimal ratio because it greatly depends on the data, task, model, and other relevant factors. Kann et al. [22] concluded that a large validation set is not practical in low-resource settings. Kann et al. [22] investigated the applicability of the validation set in low-resource NLP tasks. In this study, we empirically experiment on the validation split issue in low-resource sentence classification tasks.

Early stopping methods that are not based on a validation set can use all samples for training. This advantage is useful in low-resource settings. Choi et al. [41] proposed an early **stop**ping method **b**ased on **u**nlabeled **s**amples (BUS-stop), which performed well, particularly in low and imbalanced data settings. Garg et al. [42] also leveraged unlabeled data for early stopping but focused on providing a theoretical perspective on generalization. Bonet et al. [43] proposed a channel-wise early stopping method, which utilizes hidden features in a convolutional neural network (CNN); this method is focused on a CNN model in computer vision. Mahsereci et al. [25] proposed an evidence-based (EB) stop criterion that utilizes the gradients of the training samples. Local intrinsic dimensionality (LID) represents the minimal number of variables required to represent a part of a dataset. In few-shot image classification tasks, Lee and Chung [26] used LID as a non-validation stop criterion. However, LID and EB criteria have not been commonly used in NLP tasks. In this study, we apply them to low-resource sentence classification tasks and compare the results with two baseline stop algorithms: PE-stop-epoch and trn-conf. The details are presented in Section III. We do not consider stop methods that require unlabeled data (e.g., BUS-stop), because our study assumes that only a few labeled data are available.

## III. METHODOLOGIES

In this section, we describe the methods that are easily implementable using only a small number of labeled samples. The comparative experiments of these methods are presented in Section V.

### A. VALIDATION-BASED METHODS

The main role of a validation set is to estimate model performance, which can be used for early stopping criteria or to search for well fitted parameters. Our motivation is to investigate the efficacy of the validation set in low-resource settings.

#### 1) VAL-BASED-X%

"Val-based" refers to a general early stopping method based on the validation set. Val-based-X% denotes that X% of the labeled samples are assigned to the validation set. For example, the val-based-25% algorithm uses 25% of the labeled samples for validation and the remaining 75% for training. In our settings, the training and validation sets have the same class distribution.

#### 2) VAL-N-RUNS

The val-N-runs algorithm repeats initialization and training cycles N times, and then selects the best model based on the validation performance. Note that this is not an initialization method; however, we compare it with various initialization methods in Section V-B because the val-N-runs is a simple way to improve the performance in low-resource settings. When the proportion of the validation set is X%, we denote it as val-X%-N-runs.

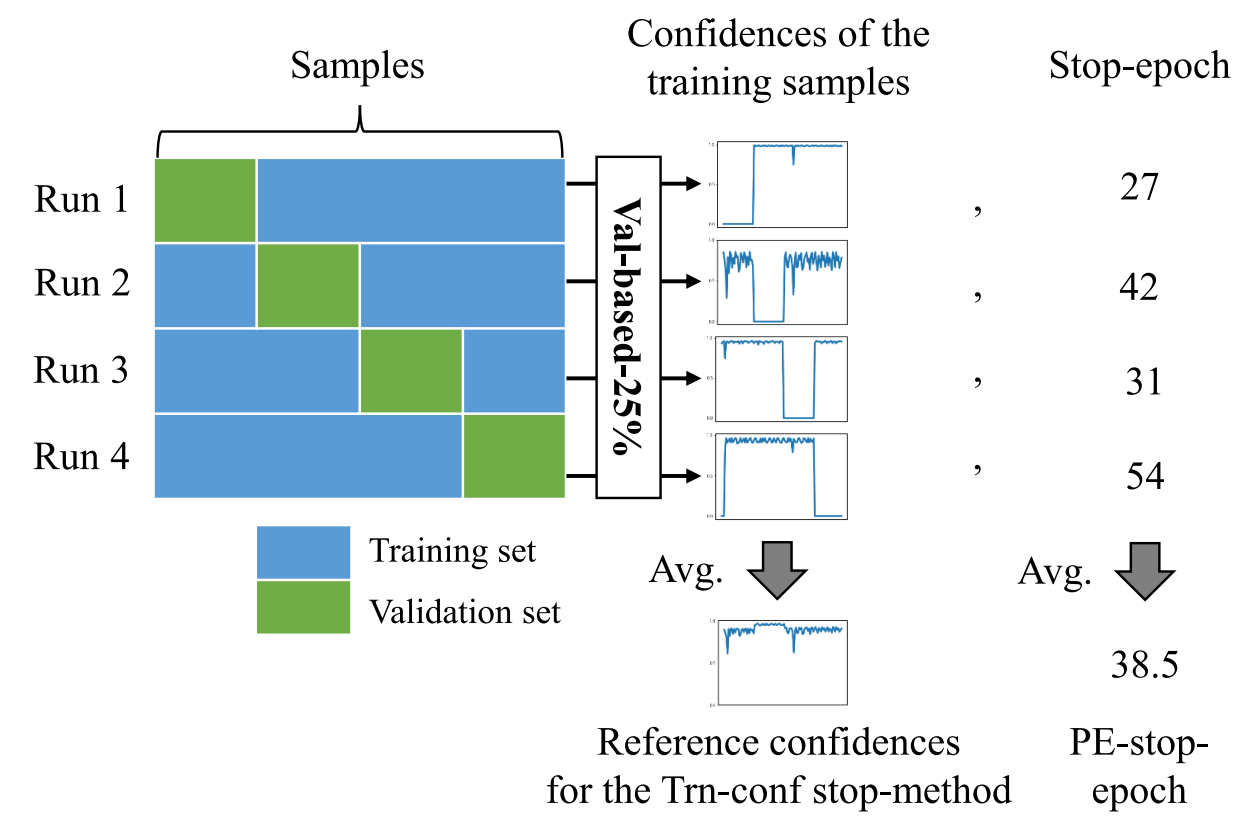

**FIGURE 1. Obtaining reference confidences and PE-stop-epoch. The reference confidences are used in the trn-conf algorithm. This example uses four-fold cross-validation.**

### B. NON-VALIDATION EARLY STOPPING CRITERIA

We introduce four non-validation stopping methods that do not require a separate validation set; thus, they can train all available samples. Our motivation is to investigate the early stopping performance of the non-validation methods and the efficacy of training all samples. Note that the trn-conf and PE-stop-epoch algorithms temporarily use the validation set to obtain stop criteria, but when subsequently training with all samples, they do not use the validation set.

#### 1) TRN-CONF

The trn-conf algorithm applies early stopping based on the confidences (i.e. probabilities associated with the predicted class label) of the training samples. First, we conduct $k$-fold cross-validation to obtain reference confidences. We use stratified cross-validation in which the class distribution is the same in the training and validation sets. In each cross-validation run, we obtain training-sample confidences when the validation performance is the best. The reference confidence is obtained by averaging the confidences of each sample. Figure 1 illustrates the process of obtaining reference confidences using four-fold cross-validation. Next, we train a new model using all samples without a validation set. The stop criterion is met at the epoch that the confidences of the training samples are most similar to the reference confidences. The confidences are sorted in the order of size before calculating the similarity. The detailed process is the same as the conf-sim method in BUS-stop [41], except that the confidences are based only on training data.

#### 2) PE-STOP-EPOCH

In the PE-stop-epoch algorithm, the stop epoch is considered a hyperparameter, which is **p**re-**e**stimated before training all labeled samples. Like the trn-conf algorithm, we conduct a stratified $k$-fold cross-validation routine in which each run stops training when the validation loss is at its lowest. Finally, the PE-stop epoch is obtained by averaging the stop epochs over $k$ runs. Figure 1 illustrates the process of obtaining the PE-stop epoch when $k$ is set to four. Thus, when training all labeled samples, we can simply stop the model at the PE-stop epoch.

#### 3) LID

Lee and Chung observed that LID can reflect the loss trend in few-shot image classification experiments [26]. LID is approximated as follows:

$$LID = -\sum_{x \in D_{trn}} \left[ \frac{1}{m} \sum_{i=1}^{m} \ln \frac{d_i(\vec{z}(x))}{d_m(\vec{z}(x))} \right]^{-1} \tag{1}$$

where $D_{trn}$ is the training set, $\vec{z}(x)$ is the representation vector of sample $x$, $d_i$ is the Euclidean distance between $\vec{z}(x)$ and its $i$-th nearest-neighbor vector, and $m$ is a hyperparameter. Differing from a previous study [26] that used both training and testing sets with LID, we apply it only to the training set $D_{trn}$. The stop criterion is met when LID is lowest.

#### 4) EB CRITERION

The EB criterion, which was proposed by Mahsereci et al. [25], is based on the local statistics of the gradients of the training samples. The optimizer halts when the gradient distribution approximates the expected value. Statistically, the stopping point implies that the gradients have difficulty carrying new information and thus the training reaches convergence. The equation is as follows [25]:

$$\frac{2}{|P|}[\log p(\nabla L_{\mathcal{S}}) - \mathbf{E}_{\nabla L_{\mathcal{S}} \sim p}[\log p(\nabla L_{\mathcal{S}})]] = 1 - \frac{|\mathcal{S}|}{|P|} \sum_{k=1}^{|P|} [\frac{(\nabla L_{\mathcal{S},k})^2}{\hat{\Sigma}_k}] > 0 \tag{2}$$

where $\mathcal{S}$ represents a subset of the training set, $P$ represents the parameters, $\nabla L$ is the loss gradient, and subscript $k$ indicates the $k$-th weight in the total parameters, $P$. $\hat{\Sigma}$ is the variance estimator, which is calculated as follows:

$$\hat{\Sigma}_k = \frac{1}{(|\mathcal{S}| - 1)} \sum_{x \in \mathcal{S}} (\nabla l_k(x) - \nabla L_{\mathcal{S},k})^{\odot 2} \tag{3}$$

where $\odot 2$ represents the elementwise square, and $\nabla l(x)$ is the gradient of loss on a sample $x$. Note that $L_{\mathcal{S}} = \frac{1}{|\mathcal{S}|} \sum_{x \in \mathcal{S}} l(x)$. As shown in the left side of Equation (2), when the computed gradient is larger than its expectation, training stops. We can use the right side of Equation (2) for its practical implementation. For more details, readers should refer to [25].

### C. METHODS OF FINDING GOOD INITIAL PARAMETERS

This section introduces several methods that can find good initialization parameters without model redesigns, which can provide crucial benefits in low-resource settings.

#### 1) NORM-INIT

In deep neural models, the truncated normal distribution (norm-init) has been widely used for initialization. It is a normal distribution whose range is restricted to $[a, b]$; we use it for baseline initialization. In our settings, mean $\mu$ and standard deviation $\sigma$ are set to 0 and 0.02, respectively, and the distribution is bounded in $[-0.04, 0.04]$.

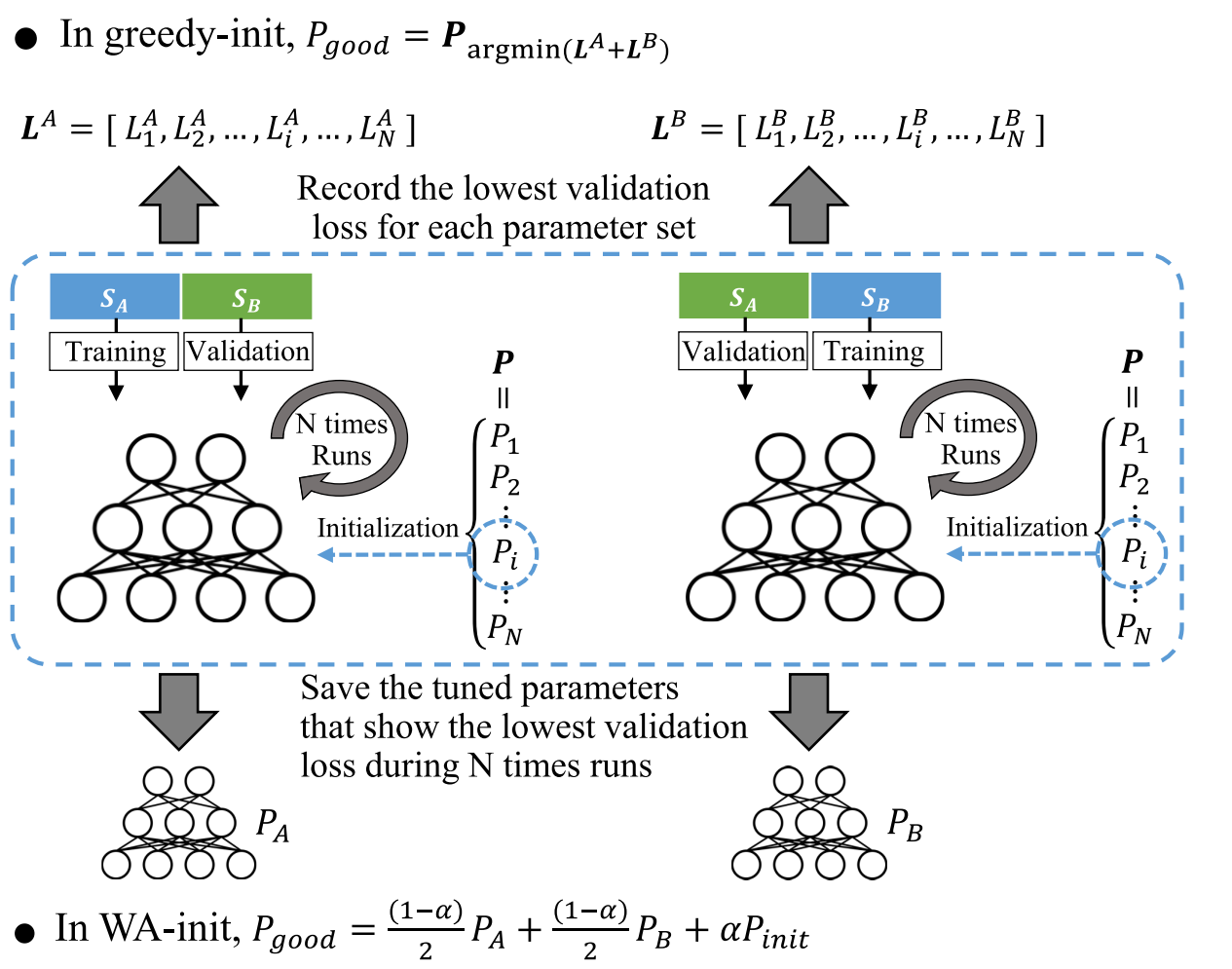

**FIGURE 2.** Our greedy-init (upper part) and WA-init (lower part) algorithms used to search for good initialization parameters. In WA-init, a certain portion of $P_i \in P$ should be initialized with the same parameters as that of $P_{init}$. Its setting is presented in Section IV-C.

### 2) GREEDY-INIT

We apply the greedy-init algorithm for multiple parameter sets $\boldsymbol{P} = \{P_1, P_2, \cdots, P_N\}$. Each set is initialized via the norm-init distribution. First, we divide the sample set $S$ into $S_A$ and $S_B$ with the same ratio. In the first phase, $S_A$ and $S_B$ are used for the training and validation sets, respectively. Next, we train the model using randomly initialized parameters $P_i$ and measure the performance on the validation set. In the second phase, we swap $S_A$ and $S_B$; thus, $S_B$ is used for training, and $S_A$ is used for validation. If the validation loss is low for both cases, then the parameters can be expected to work well with the whole sample set $S$. We thus consider the parameter set with the lowest sum of the two validation losses, which are obtained in $S_A$ and $S_B$, respectively, as good initialization parameters. The upper part of Figure 2 illustrates the greedy-init algorithm.

### 3) WA-INIT (WEIGHT AVERAGING)

Similar to the greedy-init algorithm, we split data into two subsets, $S_A$ and $S_B$. In the first phase, we conduct val-N-runs while training $S_A$ and validating $S_B$. The parameter set obtained in the first phase is denoted as $P_A$. The second phase also conducts val-N runs but trains $S_B$ and validates $S_A$. We annotate the parameter set obtained in the second phase as $P_B$. When we let the initial random (or pretrained) parameters be $P_{init}$, the good initial parameters are calculated as follows:

$$P_{good} = \frac{(1-\alpha)}{2}P_A + \frac{(1-\alpha)}{2}P_B + \alpha P_{init} \tag{4}$$

where $0 < \alpha < 1$ is a hyperparameter. As shown in Equation (4), the good initial parameters $P_{good}$ is equal to the weighted sum of $P_A$, $P_B$, and $P_{init}$. The lower part of Figure 2 illustrates the WA-init algorithm. Note that the $P_A$ (or $P_B$) is obtained by fine-tuning parameters $P_i \in \boldsymbol{P}$, and certain layers of $P_i$ should be initialized with the same parameters as those of $P_{init}$; the reason and its setting are presented in Sections VI-B and IV-C, respectively.

**TABLE 1.** Statistics for datasets. $|C|$ represents the number of classes, $|Sent|$ is the average length of sentences, and $|Data|$ is the number of samples. "Full" indicates the original setting, and "Low" indicates our low-resource setting. $D_l$ denotes the labeled set, and the training and validation sets can be split within the $D_l$. Finally, $D_{tst}$ denotes the test set.

| Dataset | $\|C\|$ | $\|Sent\|$ | Full: $\|Data\|$ | Low: $\|D_l\|/\|D_{tst}\|$ |
|---|---|---|---|---|
| CR | 2 | 20.1 | 3771 | 75 / 3696 |
| MR | 2 | 21.6 | 10662 | 213 / 10449 |
| MPQA | 2 | 3.1 | 10603 | 212 / 10391 |
| TREC | 6 | 9.8 | 5952 | 109 / 500 |
| SST-2 | 2 | 19.3 | 9613 | 155 / 1821 |
| SST-5 | 5 | 19.2 | 11855 | 192 / 2210 |

### 4) META-INIT

Dauphin and Schoenholz [28] proposed the meta-init initialization algorithm based on the assumption that the gradients of good initializations are approximately linear in the starting region. To implement this concept, they designed the gradient quotient (GQ) loss function, which is approximated as follows:

$$GQ(L, \theta) \approx \frac{1}{N}\left\| \frac{g(\theta - g(\theta))}{g(\theta) + \epsilon} - 1 \right\|_1 \tag{5}$$

where $g(\theta)$ represents the gradients of the parameters $\theta$ for loss $L$, and $\|\cdot\|_1$ is the L1-norm. A low GQ loss implies that the current gradients are similar to the gradients after one step of gradient descent. The meta-init algorithm is a data-free method; thus, the GQ loss can be trained with randomly generated data. Parameters obtained after training with GQ loss are considered as good initialization parameters.

## D. OTHER COMPARING METHODS

In addition to the methods presented above, we further compare two methods, greedy soup [34] and mixup [38]. Note that these two methods also do not require architecture redesign and any external resources.

### 1) MIXUP

For mixup, we follow the same process as described in [38]. Mixup is applied in the val-25%-10-runs manner. The loss function can be modified as follows:

$$Loss = Loss_{CE} + \frac{1}{|\mathcal{B}|}\sum_{(i,j)\sim\mathcal{S}} Loss_{mix}(x_i, y_i, x_j, y_j; \lambda) \tag{6}$$

where $\mathcal{B}$ indicates the batch set and $Loss_{CE}$ is the conventional cross-entropy loss. $\lambda$ was set to 0.5. Similar to [38], our implementation of the mixup method is trained only with $Loss_{CE}$ during the first five epochs and then together with the mixup loss after the five epochs. Readers can refer to [38] for more details.

### 2) GREEDY-SOUP

For greedy soup [34], we use 12 models trained with different hyperparameter configurations, which include learning rates of $[1e^{-5}, 2e^{-5}]$, dropout rates of [0.1, 0.2], and three random initializations for the classification layer (i.e., 2·2·3=12 fine-tuned models). Our implementation of the greedy soup algorithm uses 25% of the samples as a validation set. For more details about the algorithm, readers can refer to [34].

## IV. EXPERIMENTS

### A. DATASET AND EXPERIMENTAL SETTING

Conneau et al. [44] introduced an evaluation toolkit for English sentence representations that contains various classification datasets. These datasets are popular criteria for evaluating sentence embeddings [45]. Customer review (CR) [46], movie reviews (MR) [47], and Stanford sentiment-tree (SST) [48] datasets are sentiment analysis tasks. SST includes SST-2 and SST-5 datasets based on the number of classes. MPQA [49] is the opinion polarity classification task, and TREC [50] is the question type classification task. The statistics are presented in Table 1. In the last two columns, "Full" indicates the original dataset [44], and "Low" indicates our setting to implement the low-resource environment. $|D_l|$ is the number of labeled samples available to the training or validation sets in our low-resource setting. The split ratio for training and validation sets can be adjusted to within $|D_l|$. For CR, MR, and MPQA, 2% and 98% of the original dataset are used as the labeled set $D_l$ and test set $D_{tst}$, respectively. In case the original dataset has a standard split for training/validation/testing (e.g., TREC and SST), we use the test set as-is for testing and 2% of the original training and validation data as the labeled set. Specifically, we use 75, 213, 212, 109, 155, and 192 samples for CR, MR, MPQA, TREC, SST-2, and SST-5, respectively, as the labeled set $D_l$ as shown in Table 1. In our settings, the experiments can provide reliable results because the test set is much larger than the training set.

### B. EVALUATION METRICS

In this section, we introduce four metrics for evaluation to enable the detailed analysis.

#### 1) ACCURACY

Accuracy is a typical metric used to measure classification performance, which is defined as the rate of correctly predicted samples out of their total number. Accuracy is easy to interpret, but it does not account for the prediction probability (i.e., confidence). For example, in binary classification, two samples of 51% and 99% prediction confidence have the same contribution to accuracy. The equation is as follows:

$$Acc = \frac{1}{|\mathcal{S}|} \sum_{i \in \mathcal{S}} \mathbb{1}(y_i = \hat{y}_i) \tag{7}$$

where $\mathcal{S}$ is the set of sample indices, $y$ is the true label, $\hat{y}$ is the predicted label, and $\mathbb{1}$ is the indicator function.

#### 2) LOSS

Neural model training aims to minimize the loss function. Thus, loss can indicate how well a model is trained.

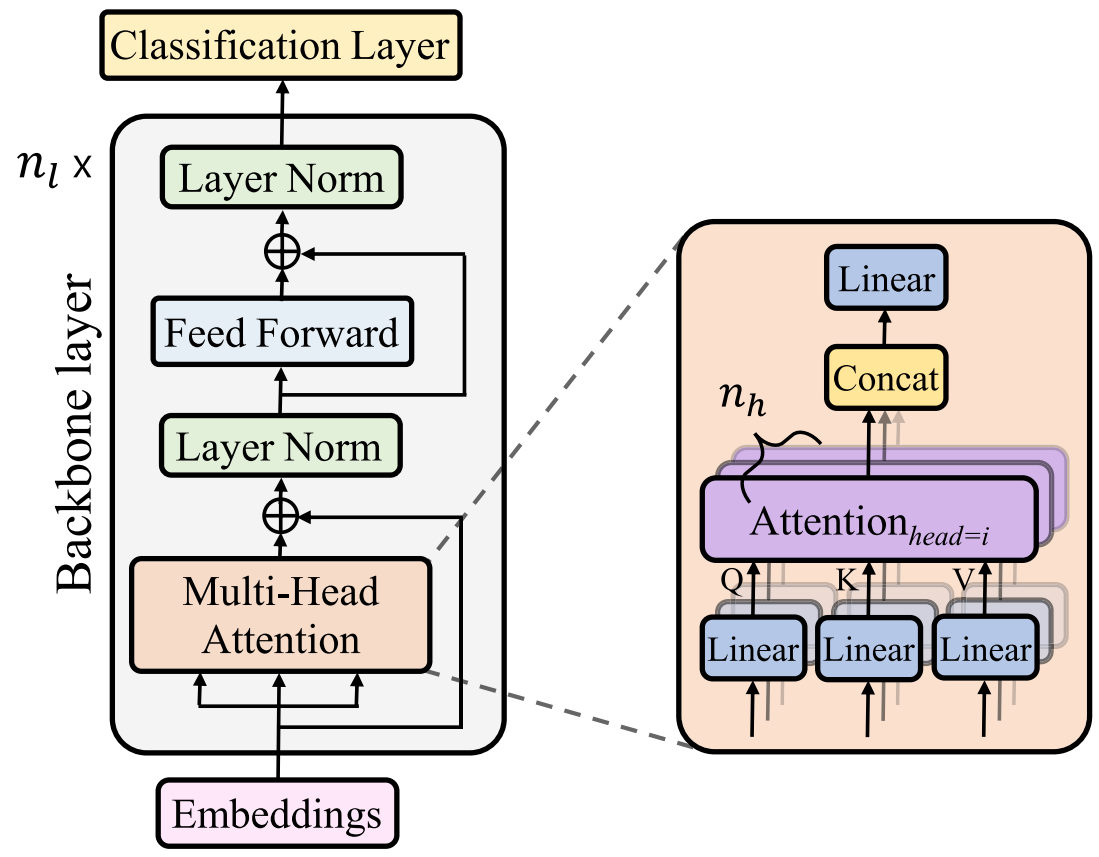

**FIGURE 3.** BERT architecture [29]. This figure is a minor modification of the model architecture in [54]. Embeddings $\in \mathbb{R}^{d \times s}$ are token vectors in a sentence of length $s$. Linear means a linear transformation, $WX + b$, where $W \in \mathbb{R}^{d_k \times d}$ and $b \in \mathbb{R}^{d_k}$. $n_h$ and $n_l$ denote the number of heads and layers, respectively. Attention block calculates Attention$(Q, K, V)$ = softmax$(QK^T/\sqrt{d_k})V$. Feed-forward is a dense layer with a dimension of $d_{ff}$ and gelu activation [55]. We use BERT-base as our default encoder. In BERT-base, $d$, $d_k$, $d_{ff}$, $n_h$, and $n_l$ are set to 768, 64, 3072, 12, and 12, respectively.

Unlike accuracy, loss can reflect the probability of the predicted label. Thus, a more confident model on the correctly predicted samples results in lower (i.e., better) loss. We use categorical cross-entropy as follows:

$$Loss_{CE} = -\frac{1}{|\mathcal{S}|} \sum_{i \in \mathcal{S}} \sum_{c \in C} \mathbb{1}(y_i = c) \log(\hat{p}_{i,c}) \tag{8}$$

where $C$ is the set of classes, and $\hat{p}_c$ is the prediction probability on class $c$.

#### 3) EXPECTED CALIBRATION ERROR

Calibration error implies that the prediction probability of each class does not always reflect the correct likelihood of the true label [27]. In real-world applications, this error is a big problem. For example, self-driving cars must consider the likelihood that an object in front of it is a pedestrian [51]. To estimate the calibration error, the predictions must be grouped into $M$ interval bins according to their confidence.

$$Acc(B_m) = \frac{1}{|B_m|} \sum_{i \in B_m} \mathbb{1}(y_i = \hat{y}_i)$$
$$Conf(B_m) = \frac{1}{|B_m|} \sum_{i \in B_m} \max_{c \in C} \hat{p}_{i,c} \tag{9}$$

where $B_m$ represents the set of sample indices whose confidence value is within the range $(\frac{m-1}{M}, \frac{m}{M}]$. The more the model is calibrated, the smaller the difference of $Conf(B_m)$ and $Acc(B_m)$ for all $m \in \{1, \cdots, M\}$. Thus, the expected calibration error (ECE) is defined as follows [27]:

$$ECE = \sum_{m=1}^{M} \frac{|B_m|}{|\mathcal{S}|} \left| Acc(B_m) - Conf(B_m) \right| \tag{10}$$

The ECE measures the level of model calibration by considering both overconfident and underconfident samples.

#### 4) OVERCONFIDENCE ERROR

Recent studies have observed that deep neural networks tend to be overconfident when trained with hard labels [52], [53]. Therefore, we consider a metric to estimate the overconfidence error (OE). Thulasidasan et al. defined an OE as follows [52]:

$$OE = \sum_{m=1}^{M} \frac{|B_m|}{|S|} \Big[ Conf(B_m) \times \max(Conf(B_m) - Acc(B_m), 0) \Big] \quad (11)$$

Unlike the ECE, the OE considers only overconfident samples. In this study, we use ECE and OE to estimate calibration performance.

### C. IMPLEMENTATION DETAILS

For evaluation, we apply BERT-base [29] as our base text encoder. All the methods in this study, including early stopping and initialization methods, are implemented using the BERT-base model. BERT architecture and its configuration are illustrated in Figure 3. For the training hyperparameters, we follow the default settings presented in [29] instead of newly tuning on a validation set, because small validation sets are prone to producing inappropriate hyperparameters. The experiment in Section VI-C supports this fact. The Adam optimizer [56] is applied with a learning rate of $2e^{-5}$ and a dropout rate of 0.1. The batch size is 32, and the maximum sentence length is set to 64. The word embeddings are not updated because in low-resource settings, many words in the test set do not appear in the training set. For the trn-conf and PE-stop-epoch algorithms, we use four-fold cross-validation. For LID, $m$ is set to 64. For the EB criterion, we use 64 random training samples for the sample set $\mathcal{S}$. The norm-init, greedy-init, and meta-init methods are applied only to the classification layer because BERT-base is a pre-trained model. For the meta-init algorithm, we use the code provided by Dauphin and Schoenholz [28]. For all $P_i \in \boldsymbol{P}$ in the WA-init algorithm, their backbone layers are initialized with the same BERT pretrained parameters while the classification layers are initialized with different parameters (see Figures 2 and 3). In Equation (4), $\alpha$ is set to 1/3. The experiments are conducted in Python 3.6 on a computer having a 3.50-GHz Intel Core i7-5930K processor, 64 GB RAM, and an NVIDIA Titan Xp GPU with 12 GB memory, using Tensorflow 1.15.0 library [57].

## V. RESULTS

### A. COMPARISON OF STOP CRITERIA

In this section, we compare the early stopping criteria in our low-resource settings. Table 2 lists the results of different stop criteria for each dataset, and Table 3 displays the average result over all datasets. Note that val-based-X% algorithms use some fractions (X%) of the labeled samples for validation, whereas the non-validation stopping methods (i.e., trn-conf, EB criterion, LID, and PE-stop-epoch) use all labeled samples for training. In the val-based-X% algorithm, the ratio of the validation set X% was set to 75%, 50%, and 25%, respectively. As shown in Table 3, validation-based stopping generally resulted in lower accuracy but better calibration performance than the non-validation stop methods; this trend was more pronounced as the ratio of the validation set increased. The larger validation set can represent the test set more adequately, thereby reducing the calibration error. However, prediction performance (i.e., accuracy and loss) was low because the large validation set reduces the training samples by that amount. This result suggests that in low-resource settings, even a small amount of data have a great impact depending on whether it is used for training or validation. It is also noted that the primary evaluation metric should be prediction performance (i.e., accuracy and loss), whereas the calibration error should be used as a secondary metric. For example, if all the samples were predicted as 51% confidence and the accuracy was equally 51%, the calibration error would be zero. In this context, validation-based stopping that showed low prediction performance may not be the best for low-resource tasks. Meanwhile, the EB criterion resulted in the best accuracy on average, whereas it showed relatively low performance in terms of loss and calibration error. The PE-stop-epoch algorithm showed the second-best accuracy and third-best loss, but it also showed a high calibration error compared with validation-based stopping. Unlike other non-validation stop methods, the trn-conf algorithm resulted in low calibration errors. As shown in Table 3, the ECE and OE of the trn-conf algorithm were as low as 44% and 53% of the EB criterion, respectively. The trn-conf algorithm showed balanced performances across the four evaluation metrics; it showed the lowest loss in most datasets and its accuracy was comparable to the EB criterion on average.

The overall results show that the performance can significantly differ, depending on the stop criteria and evaluation metric. In Table 3, the maximum difference in absolute accuracy was 14.8% as calculated between the val-based-75% algorithm and EB criterion. The loss differed by up to 35.5%, which falls between the trn-conf algorithm and LID, when letting the best performance be 100%. The ECE and OE also differed by up to 58.8% and 71.2%, respectively. Researchers can thus select a suitable stop method while considering different evaluation metric performances. Section VI-A analyzes why each stop method showed different results in the evaluation metrics.

### B. COMPARISON OF GOOD PARAMETERS

In this section, we compare various methods (i.e., norm-init, greedy-init, WA-init, and meta-init methods for good initialization, and mixup and greedy-soup for other learning techniques) that can be applied using only a small amount of labeled data without architectural redesign. We experimented with the val-X%-10-runs algorithm as a baseline. Note that the given initialization methods train all labeled samples, whereas the val-X%-10-runs, mixup, and greedy-soup algorithms use validation sets of size X%, 25%, and

**TABLE 2.** Performances of different stop criteria on various evaluation metrics. Each model used only 2% of the total samples (75 to 213) for each dataset. The best performances are shown in bold. Max. Diff. denotes the maximum performance difference between stop criteria. Note that stop criteria trn-conf, EB, LID, and PE-stop-epoch train all labeled samples, whereas val-based-X% methods train (100-X)% of the labeled samples.

| Dataset | CR | | | | MR | | | | TREC | | | |
|---|---|---|---|---|---|---|---|---|---|---|---|---|
| Method | Acc | Loss | ECE | OE | Acc | Loss | ECE | OE | Acc | Loss | ECE | OE |
| Val-based-75% | 0.681±.044 | 0.602±.049 | **0.069**±.036 | **0.042**±.033 | 0.671±.044 | 0.617±.039 | 0.059±.033 | 0.039±.033 | 0.591±.078 | 1.133±.176 | 0.112±.050 | 0.056±.046 |
| Val-based-50% | 0.722±.056 | 0.555±.076 | 0.071±.032 | 0.043±.033 | 0.747±.040 | 0.532±.052 | **0.056**±.027 | **0.038**±.027 | 0.729±.054 | 0.822±.135 | 0.091±.036 | 0.044±.036 |
| Val-based-25% | 0.784±.049 | **0.498**±.099 | 0.073±.040 | 0.050±.045 | 0.768±.030 | 0.515±.046 | 0.067±.033 | 0.046±.038 | 0.791±.087 | 0.639±.168 | 0.078±.031 | **0.039**±.029 |
| Trn-conf | 0.795±.045 | 0.512±.120 | 0.092±.040 | 0.077±.042 | 0.780±.023 | **0.507**±.049 | 0.076±.031 | 0.061±.031 | 0.828±.039 | 0.562±.124 | **0.068**±.028 | **0.039**±.031 |
| EB criterion | 0.804±.038 | 0.565±.114 | 0.126±.034 | 0.116±.033 | **0.793**±.019 | 0.649±.064 | 0.150±.017 | 0.141±.016 | 0.830±.041 | **0.551**±.134 | 0.069±.031 | 0.047±.029 |
| LID | **0.810**±.039 | 0.803±.245 | 0.147±.049 | 0.139±.053 | 0.741±.058 | 0.615±.150 | 0.097±.048 | 0.047±.063 | 0.755±.164 | 0.803±.381 | 0.128±.070 | 0.074±.048 |
| PE-stop-epoch | 0.806±.040 | 0.618±.142 | 0.133±.038 | 0.123±.038 | 0.784±.024 | 0.556±.080 | 0.113±.031 | 0.100±.031 | **0.832**±.038 | 0.574±.144 | 0.084±.034 | 0.067±.034 |
| Max. Diff. | 0.125 | 0.305 | 0.078 | 0.097 | 0.122 | 0.142 | 0.094 | 0.103 | 0.241 | 0.582 | 0.044 | 0.035 |
| **Dataset** | MPQA | | | | SST-2 | | | | SST-5 | | | |
| **Method** | Acc | Loss | ECE | OE | Acc | Loss | ECE | OE | Acc | Loss | ECE | OE |
| Val-based-75% | 0.718±.033 | 0.566±.040 | **0.065**±.026 | **0.041**±.026 | 0.694±.067 | 0.585±.078 | **0.067**±.030 | **0.040**±.032 | 0.276±.041 | 1.583±.027 | **0.052**±.036 | **0.018**±.019 |
| Val-based-50% | 0.789±.046 | 0.484±.068 | 0.067±.027 | 0.048±.029 | 0.786±.043 | 0.483±.060 | **0.067**±.028 | 0.042±.032 | 0.326±.048 | 1.546±.048 | 0.076±.053 | 0.037±.037 |
| Val-based-25% | 0.830±.041 | 0.441±.079 | 0.069±.030 | 0.053±.036 | 0.816±.044 | 0.455±.081 | 0.073±.027 | 0.050±.037 | 0.364±.052 | 1.481±.080 | 0.086±.054 | 0.045±.040 |
| Trn-conf | 0.847±.019 | **0.409**±.051 | **0.065**±.025 | 0.054±.027 | 0.828±.035 | **0.433**±.086 | 0.077±.032 | 0.061±.033 | 0.391±.031 | **1.468**±.096 | 0.121±.041 | 0.069±.029 |
| EB criterion | 0.850±.028 | 0.501±.065 | 0.107±.017 | 0.100±.021 | 0.837±.026 | 0.497±.083 | 0.112±.024 | 0.103±.023 | **0.404**±.023 | 1.963±.133 | 0.382±.027 | 0.319±.028 |
| LID | 0.779±.072 | 0.555±.134 | 0.090±.036 | 0.054±.048 | 0.783±.063 | 0.508±.106 | 0.090±.036 | 0.029±.038 | 0.403±.023 | 2.747±.169 | 0.480±.022 | 0.439±.022 |
| PE-stop-epoch | **0.852**±.017 | 0.495±.100 | 0.096±.029 | 0.089±.032 | **0.838**±.024 | 0.496±.103 | 0.105±.026 | 0.096±.027 | 0.394±.031 | 1.510±.110 | 0.183±.054 | 0.119±.044 |
| Max. Diff. | 0.134 | 0.157 | 0.042 | 0.059 | 0.144 | 0.152 | 0.045 | 0.063 | 0.128 | 1.279 | 0.428 | 0.421 |

**TABLE 3.** Average performances of stop criteria over all datasets. (·) indicates a relative percentage when letting the best performance among the methods be 100%.

| Dataset | Average | | | |
|---|---|---|---|---|
| Method | Acc | Loss | ECE | OE |
| Val-based-75% | 0.605 (80.4) | 0.848 (76.5) | **0.071** (100.) | **0.040** (100.) |
| Val-based-50% | 0.683 (90.7) | 0.737 (88.0) | 0.071 (99.2) | 0.042 (94.1) |
| Val-based-25% | 0.726 (96.4) | 0.672 (96.6) | 0.075 (94.9) | 0.047 (84.0) |
| Trn-conf | 0.745 (98.9) | **0.649** (100.) | 0.083 (85.3) | 0.060 (65.7) |
| EB criterion | **0.753** (100.) | 0.787 (82.4) | 0.158 (44.9) | 0.137 (28.8) |
| LID | 0.712 (94.5) | 1.005 (64.5) | 0.172 (41.2) | 0.130 (30.4) |
| PE-stop-epoch | 0.751 (99.7) | 0.708 (91.6) | 0.119 (59.4) | 0.099 (40.0) |
| Max. Diff | 0.148 (19.6) | 0.356 (35.5) | 0.101 (58.8) | 0.097 (71.2) |

25%, respectively. For the initialization methods, we applied the trn-conf algorithm as a stop criterion because it showed good balanced performance across the four evaluation metrics. In the greedy-init and WA-init algorithms, the hyperparameter N was set to 5 so that the training times would be the same as that in the val-X%-10-runs algorithm. Table 4 lists the performance results for each dataset, and Table 5 lists the average performances over all datasets. In most datasets, the WA-init algorithm showed the best accuracy and loss. On average, the val-75%-10-runs algorithm showed the best ECE. However, this was caused by the low prediction performance discussed in the previous section. As shown in Table 5, the prediction performance increased as the ratio of the validation set was reduced (i.e., more samples were trained). As such, in a low-resource setting, a large validation set may not be effective. The greedy-soup algorithm showed the best OE but the second worst ECE. The simple average of multiple models only prevented overconfident predictions while, overall, distorting the likelihood of the correct labels. Further, the prediction performances of the mixup and val-25%-10-runs algorithms were statistically similar and slightly higher than those of the greedy-soup algorithm. Note that the mixup and greedy-soup algorithms also use a validation set of 25% size. We speculate that the simple val-N-runs algorithm with a sufficiently high N can encompass the improvement by the mixup and greedy-soup algorithms. Meanwhile, the meta-init algorithm showed low performance overall. We speculate that this occurred because the meta-init algorithm, a data-free method, searches for good initial parameters without reflecting the data features. We also observed that, on average, the val-25%-10-runs algorithm performed slightly better than the norm-init method, which trains all samples. This result indicates that a validation set in a low-resource setting can be effectively used for performance improvement. Meanwhile, the prediction performances of the greedy-init and WA-init algorithms were consistently better than those of the val-25%-10-runs algorithm in all datasets. Table 5 shows that the WA-init algorithm improved the average accuracy by 2.0% from the normally initialized parameters (i.e., norm-init). The val-X%-N-runs, norm-init, greedy-init, and WA-init methods differ only in how they exploit the validation set. Note that the greedy-init and WA-init algorithms temporarily use the validation set to find good initial parameters before training all samples. This implies that performance can vary, depending on how the samples are exploited in terms of the training-validation split and initialization. As shown in Tables 4 and 5, the performance differences between methods were significantly large. For example, in Table 5, the maximum difference in absolute accuracy was more than 10%. Even when using the same data and architecture, the training strategy can be very effective in low-resource settings.

**TABLE 4.** Performances of different initialization, mixup, greedy-soup, and val-X%-10-runs algorithms on various evaluation metrics. Each model used only 2% of the total samples (75 to 213) for each dataset. The best performances are shown in bold. Max. Diff. denotes the maximum difference in performance. Note that initialization methods norm-init, greedy-init, WA-init, and meta-init train all labeled samples, whereas the val-X%-N-runs methods train (100-X)% of the labeled samples, and mixup and greedy-soup algorithms train 75% of the labeled samples.

| Dataset | CR | | | | MR | | | | TREC | | | |
|---|---|---|---|---|---|---|---|---|---|---|---|---|
| Method | Acc | Loss | ECE | OE | Acc | Loss | ECE | OE | Acc | Loss | ECE | OE |
| Val-75%-10-runs | 0.719±.048 | 0.566±.063 | **0.068**±.034 | 0.049±.032 | 0.721±.040 | 0.565±.046 | **0.057**±.026 | 0.040±.027 | 0.635±.072 | 1.019±.162 | 0.093±.032 | 0.044±.033 |
| Val-50%-10-runs | 0.776±.056 | 0.499±.088 | 0.071±.037 | 0.055±.036 | 0.772±.027 | 0.503±.039 | 0.058±.027 | 0.043±.028 | 0.758±.049 | 0.736±.127 | 0.087±.034 | 0.044±.035 |
| Val-25%-10-runs | 0.806±.041 | 0.510±.141 | 0.083±.050 | 0.067±.055 | 0.786±.019 | 0.503±.059 | 0.068±.036 | 0.050±.040 | 0.805±.041 | 0.626±.125 | 0.084±.045 | 0.044±.037 |
| Mixup | 0.802±.040 | 0.570±.207 | 0.097±.046 | 0.084±.049 | 0.785±.019 | 0.504±.055 | 0.071±.038 | 0.052±.041 | 0.814±.039 | 0.581±.103 | 0.075±.028 | 0.037±.027 |
| Greedy-soup | 0.779±.052 | 0.539±.075 | 0.127±.064 | **0.027**±.039 | 0.776±.029 | 0.544±.060 | 0.110±.051 | **0.033**±.045 | 0.790±.056 | 0.839±.268 | 0.215±.142 | **0.012**±.021 |
| Norm-init | 0.795±.045 | 0.512±.120 | 0.092±.040 | 0.077±.042 | 0.780±.023 | 0.507±.049 | 0.076±.031 | 0.061±.031 | 0.828±.039 | 0.562±.124 | 0.068±.028 | 0.039±.031 |
| Greedy-init | 0.820±.042 | 0.452±.104 | 0.071±.037 | 0.057±.038 | 0.788±.029 | 0.495±.054 | 0.073±.031 | 0.059±.030 | 0.834±.039 | **0.543**±.122 | **0.066**±.023 | 0.038±.026 |
| WA-init | **0.831**±.040 | **0.437**±.098 | 0.070±.030 | 0.057±.033 | **0.796**±.021 | **0.490**±.042 | 0.076±.023 | 0.061±.024 | **0.839**±.038 | 0.559±.123 | 0.072±.030 | 0.047±.032 |
| Meta-init | 0.714±.086 | 1.723±1.788 | 0.211±.111 | 0.199±.111 | 0.724±.082 | 1.454±2.101 | 0.187±.106 | 0.175±.107 | 0.725±.136 | 1.211±1.334 | 0.135±.135 | 0.108±.137 |
| Max. Diff. | 0.117 | 1.286 | 0.143 | 0.150 | 0.075 | 0.964 | 0.130 | 0.135 | 0.204 | 0.668 | 0.069 | 0.070 |

| Dataset | MPQA | | | | SST-2 | | | | SST-5 | | | |
|---|---|---|---|---|---|---|---|---|---|---|---|---|
| Method | Acc | Loss | ECE | OE | Acc | Loss | ECE | OE | Acc | Loss | ECE | OE |
| Val-75%-10-runs | 0.768±.043 | 0.502±.061 | 0.061±.026 | 0.043±.027 | 0.770±.046 | 0.497±.068 | 0.060±.027 | 0.039±.030 | 0.316±.031 | 1.545±.035 | **0.057**±.035 | **0.023**±.020 |
| Val-50%-10-runs | 0.830±.031 | 0.422±.066 | 0.057±.029 | 0.045±.031 | 0.823±.028 | 0.419±.044 | **0.055**±.016 | 0.034±.022 | 0.366±.027 | 1.484±.054 | 0.087±.050 | 0.046±.034 |
| Val-25%-10-runs | 0.852±.019 | 0.438±.095 | 0.073±.036 | 0.059±.042 | 0.839±.017 | 0.419±.066 | 0.072±.027 | 0.053±.034 | 0.396±.024 | 1.419±.060 | 0.095±.048 | 0.053±.035 |
| Mixup | 0.855±.024 | 0.430±.149 | 0.061±.037 | 0.050±.042 | 0.841±.020 | 0.438±.174 | 0.069±.026 | 0.051±.035 | 0.391±.027 | 1.428±.057 | 0.119±.048 | 0.070±.036 |
| Greedy-soup | 0.854±.023 | 0.438±.071 | 0.113±.069 | **0.028**±.038 | 0.834±.026 | 0.471±.071 | 0.130±.069 | **0.025**±.036 | 0.384±.032 | 1.461±.079 | 0.099±.064 | 0.046±.053 |
| Norm-init | 0.847±.019 | 0.409±.051 | 0.065±.025 | 0.054±.027 | 0.828±.035 | 0.433±.086 | 0.077±.032 | 0.061±.033 | 0.391±.031 | 1.468±.096 | 0.121±.041 | 0.069±.029 |
| Greedy-init | 0.860±.017 | 0.377±.045 | 0.054±.020 | 0.045±.022 | 0.847±.021 | **0.392**±.051 | 0.060±.023 | 0.045±.025 | 0.401±.023 | 1.413±.050 | 0.120±.039 | 0.070±.028 |
| WA-init | **0.864**±.015 | **0.368**±.038 | **0.053**±.019 | 0.041±.021 | **0.854**±.013 | 0.399±.037 | 0.069±.020 | 0.055±.023 | **0.408**±.025 | **1.379**±.046 | 0.102±.035 | 0.057±.024 |
| Meta-init | 0.772±.089 | 1.153±1.281 | 0.164±.103 | 0.154±.104 | 0.754±.104 | 1.317±1.017 | 0.188±.123 | 0.178±.123 | 0.300±.072 | 2.389±.974 | 0.292±.152 | 0.202±.129 |
| Max. Diff. | 0.096 | 0.785 | 0.111 | 0.113 | 0.100 | 0.925 | 0.133 | 0.144 | 0.108 | 1.010 | 0.235 | 0.179 |

**TABLE 5.** Average performances of the initialization and val-X%-10-runs algorithms over all datasets. (·) indicates the relative percentage when letting the best performance among methods be 100%.

| Dataset | Average | | | |
|---|---|---|---|---|
| Method | Acc | Loss | ECE | OE |
| Val-75%-10-runs | 0.655 (85.6) | 0.782 (77.4) | **0.066** (100.) | 0.040 (73.4) |
| Val-50%-10-runs | 0.721 (94.2) | 0.677 (89.4) | 0.069 (95.6) | 0.044 (65.9) |
| Val-25%-10-runs | 0.747 (97.7) | 0.653 (92.7) | 0.079 (83.5) | 0.054 (53.7) |
| Mixup | 0.748 (97.8) | 0.658 (91.9) | 0.082 (80.5) | 0.057 (50.9) |
| Greedy-soup | 0.736 (96.2) | 0.715 (84.6) | 0.132 (50.0) | **0.029** (100.) |
| Norm-init | 0.745 (97.3) | 0.649 (93.3) | 0.083 (79.7) | 0.060 (48.3) |
| Greedy-init | 0.758 (99.1) | 0.612 (98.9) | 0.074 (89.3) | 0.052 (55.8) |
| WA-init | **0.765** (100.) | **0.605** (100.) | 0.073 (90.0) | 0.053 (54.7) |
| Meta-init | 0.665 (86.9) | 1.541 (39.2) | 0.196 (33.7) | 0.169 (17.2) |
| Max. Diff. | 0.110 (14.4) | 0.936 (60.8) | 0.130 (66.3) | 0.129 (76.6) |

**TABLE 6.** Accuracy when applying our approaches to PERFECT model [58]. The average accuracies on the datasets of methods val-based-25%, val-25%-10-runs, norm-init, and WA-init are 0.7137, 0.7202, 0.7435, and 0.7482, respectively. WA-init shows the best average performance in the PERFECT model as well.

| *Results applied to PERFECT model* | | | | | | |
|---|---|---|---|---|---|---|
| Method | CR | MR | TREC | MPQA | SST-2 | SST-5 |
| Val-based-25% | 0.744 | 0.785 | 0.706 | **0.859** | 0.804 | 0.384 |
| Val-25%-10-runs | **0.756** | **0.790** | **0.725** | 0.858 | **0.806** | **0.386** |
| Improvement (%) | 1.2%↑ | 0.5%↑ | 1.9%↑ | 0.1%↓ | 0.2%↑ | 0.2%↑ |
| Norm-init | 0.798 | **0.802** | 0.754 | **0.866** | 0.829 | 0.412 |
| WA-init | **0.804** | 0.801 | **0.761** | 0.865 | **0.842** | **0.416** |
| Improvement (%) | 0.6%↑ | 0.1%↓ | 0.7%↑ | 0.1%↓ | 1.3%↑ | 0.4%↑ |

## C. ADAPTATION TO OTHER METHODS

Our integrated method, WA-init, is shown to work well even with small amounts of labeled data. Its advantage is that it can be easily extended to other low-resource approaches. To investigate its applicability, we adapted WA-init to other three recent SOTA methods: (1) PERFECT, (2) UST, and (3) LAMBADA; they are model redesign, self-learning, and data augmentation techniques, respectively. Note that these methods are not directly comparable with our approach because they employ external resources or structurally enhanced models. Therefore, we focus on how our approach WA-init can bring additional performance improvement when adapted to these methods.

### 1) PERFECT

PERFECT stands for Prompt-free and Efficient paRadigm for FEw-shot Cloze-based fine-Tuning [58]. The PERFECT model achieved SOTA performance on various NLP tasks, taking advantage of the pre-trained language model. However, PERFECT model requires architecture redesign; it inserts several hidden layers between existing blocks. We applied WA-init to the PERFECT model using the BERT architecture. Hyperparameters for the PERFECT model, the number of mask tokens and learning rate, were selected based on validation performance. Specifically, the number of mask tokens were 2 for all datasets and the learning rates were $1e^{-3}$, $1e^{-3}$, $1e^{-5}$, $1e^{-4}$, $1e^{-4}$, and $1e^{-4}$ for CR, MR, TREC, MPQA, SST-2, and SST-5, respectively. When applying WA-init, we used the same good initial parameters $P_g$ as those used in Section V-B (i.e., $P_g$ is obtained using BERT and thereafter used to initialize the BERT layers within the PERFECT model). In this experiment, $\alpha$ in Equation (4) was set to a higher value of 0.8 because the PERFECT model highly relies on the initial parameters of the pre-trained language model. We also reduced the data usage to 1% of the total to clearly reveal the performance difference between the algorithms. Four algorithms were implemented on the PERFECT model: val-based-25%, val-25%-N-runs, norm-init, and WA-init. PE-stop-epoch was used as the stop criterion. Table 6 presents the resulting accuracy. The results are similar to those in Section V-B. The val-25%-10-runs algorithm exhibits better overall performance than the val-based-25% algorithm (i.e., val-25%-1-runs). Despite the lack of data, the result demonstrates that high validation performance generally leads to a high test accuracy without serious overfitting. In addition, training all samples with norm-init or WA-init consistently yielded better results than the validation-based methods. Further, the WA-init improved the average performance over the normal initial parameters (i.e., norm-init), demonstrating its applicability in the SOTA model. This result indicates that our approach can be effectively adapted to other low-resource models.

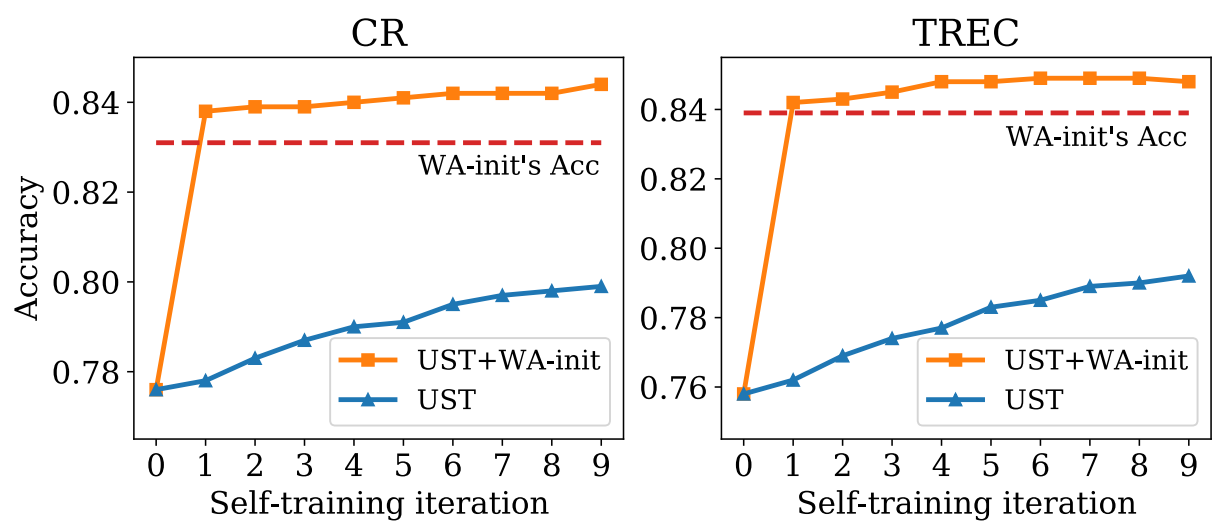

**FIGURE 4.** Accuracy during self-training iterations. UST [59] is used as the self-training algorithm. When WA-init was applied to UST, the overall accuracy improved.

**TABLE 7.** Final accuracy after 10 self-training iterations. UST consistently improves the performance for all datasets. Further, when adapting WA-init into UST, the performance improved for all datasets (UST→UST+WA-init). The average accuracy on the datasets of methods Val-50%-10-runs, UST, WA-init, and UST+WA-init are 0.7208, 0.7365, 0.7653, and 0.7738, respectively.

| *Results applied to self-training* | | | | | | |
|---|---|---|---|---|---|---|
| **Method** | CR | MR | TREC | MPQA | SST-2 | SST-5 |
| Val-50%-10-runs | 0.776 | 0.772 | 0.758 | 0.830 | 0.823 | 0.366 |
| +UST | **0.799** | **0.776** | **0.792** | **0.846** | **0.835** | **0.371** |
| Improvement (%) | 2.3%↑ | 0.4%↑ | 3.4%↑ | 1.6%↑ | 1.2%↑ | 0.5%↑ |
| WA-init | 0.831 | 0.796 | 0.839 | 0.864 | 0.854 | 0.408 |
| UST+WA-init | **0.844** | **0.804** | **0.848** | **0.873** | **0.859** | **0.415** |
| Improvement (%) | 1.3%↑ | 0.8%↑ | 0.9%↑ | 0.9%↑ | 0.5%↑ | 0.7%↑ |

### 2) UST

UST stands for Uncertainty-aware Self-Training [59]; it showed SOTA performance in few-shot text classification by leveraging unlabeled data. UST calculates the uncertainty of the prediction of unlabeled data and utilizes it in an iterative self-training mechanism. In this experiment, we investigate how our approach can help in self-training. We re-implement UST in our experimental settings. We compare the following four model settings: val-50%-10-runs, UST, WA-init, and UST+WA-init. The settings val-50%-10-runs and WA-init are the same as those presented in Section V-B. UST follows the same process as in [59]. The setting UST+WA-init has the same process as the UST; however, only at the first iteration, the pseudo labels of unlabeled data are obtained by the model trained with WA-init. All the four settings use 75–213 labeled data (i.e., $|D_l|$ in Table 1) with 2,000 additional unlabeled data for each dataset. In UST and UST+WA-init, the self-training iteration starts from the trained val-50%-10-runs model. Figure 4 shows the accuracy at every iteration. UST+WA-init drastically increases the accuracy at the first iteration and consistently outperforms the normal UST during all iterations. Indeed, WA-init can help iterative self-training by providing better prediction results in the initial step. Table 7 shows the average accuracy for each dataset. Overall, UST+WA-init outperforms the UST and WA-init each. This result indicates that our approach can bring additional performance gain in a self-training mechanism. Meanwhile, WA-init, which uses all of the small amount of labeled data, significantly outperforms the UST, which additionally uses 2,000 unlabeled data with half of the labeled data. This result again emphasizes the importance of well exploiting given labeled data.

**TABLE 8.** Accuracy when a data augmentation method, LAMBADA [60], is applied to norm-init and WA-init. The average accuracies on the datasets of norm-init, norm-init+LAMBADA, WA-init, and WA-init+LAMBADA are 0.7448, 0.7528, 0.7653, and 0.7648, respectively. The WA-init algorithm without LAMBADA applied shows the best performance on average.

| *Results applied to data augmentation* | | | | | | |
|---|---|---|---|---|---|---|
| **Method** | CR | MR | TREC | MPQA | SST-2 | SST-5 |
| Norm-init | 0.795 | 0.780 | **0.828** | 0.847 | 0.828 | 0.391 |
| +LAMBADA | **0.812** | **0.793** | 0.821 | **0.852** | **0.844** | **0.395** |
| Improvement (%) | 1.7%↑ | 1.3%↑ | 0.7%↓ | 0.5%↑ | 1.6%↑ | 0.4%↑ |
| WA-init | **0.831** | 0.796 | **0.839** | **0.864** | 0.854 | 0.408 |
| +LAMBADA | 0.825 | **0.807** | 0.830 | 0.859 | 0.854 | **0.414** |
| Improvement (%) | 0.6%↓ | 1.1%↑ | 0.9%↓ | 0.5%↓ | 0.0%- | 0.6%↑ |

### 3) LAMBADA

LAMBADA stands for LAnguage-Model-BAsed Data Augmentation, which is a SOTA data augmentation method for text classification [60]. LAMBADA employs a large pre-trained language model, GPT-2 (Generative Pre-Training) [61], to generate new training data. We reproduced LAMBADA as described in [60]. We used `gpt2-large` model provided in the Huggingface library [62]. All the hyperparameters and training process of the `gpt2-large` model follow [61]. For each dataset, we generated the same amount of data as the given labeled set (i.e., $|D_l|$ in Table 1); thus, training data were doubled. Using the augmented data, we experimented with two initializations: norm-init and WA-init. Note that both existing labeled samples (i.e., 2% of the original dataset) and generated samples are trained.

**TABLE 9. Improved accuracy when increasing the number of runs N and sample splits |𝕊| in WA-init algorithm. The average accuracy on the datasets of WA-init (N = 5, |𝕊| = 2) and WA-init (N = 10, |𝕊| = 4) are 0.7653 and 0.7703, respectively.**

*Results of adjusting N and |𝕊| in WA-init*

| Method | CR | MR | TREC | MPQA | SST-2 | SST-5 |
|---|---|---|---|---|---|---|
| WA-init (N=5, \|𝕊\|=2) | 0.831 | 0.796 | 0.839 | 0.864 | **0.854** | 0.408 |
| WA-init (N=10, \|𝕊\|=4) | **0.840** | **0.804** | **0.845** | **0.865** | 0.853 | **0.415** |
| Improvement (%) | 0.9%↑ | 0.8%↑ | 0.6%↑ | 0.1%↑ | 0.1%↓ | 0.7%↑ |

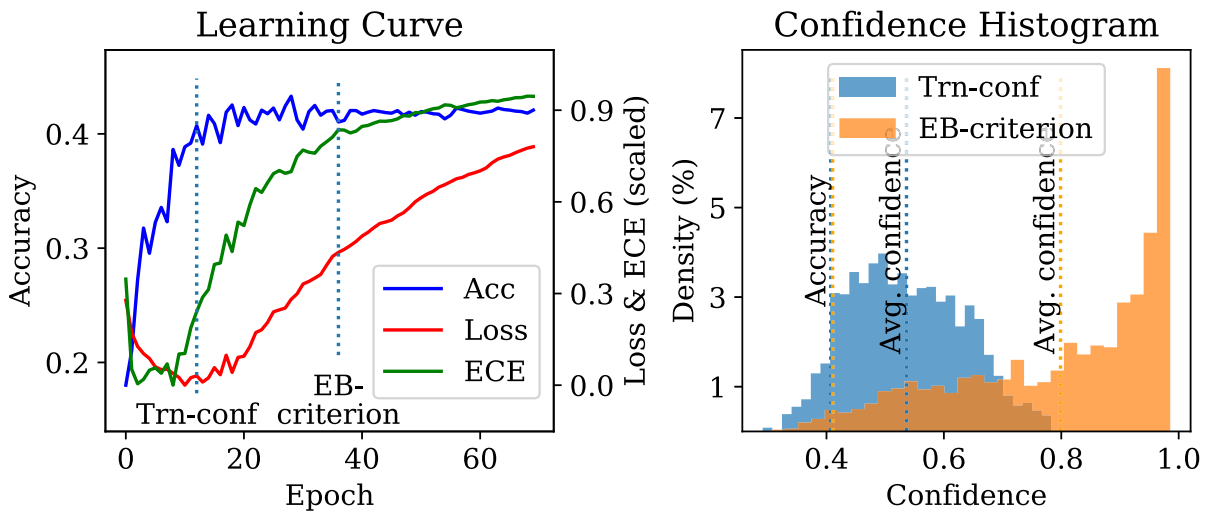

**FIGURE 5. Comparison of two early stopping criteria: trn-conf and EB. The left side displays the accuracy, loss, and ECE curves of the SST-5 test set, and the right side displays the confidence histograms of the test data when stopped with the trn-conf algorithm and EB criterion, respectively.**

Table 8 presents the results. When LAMBADA-generated data were added to norm-init, the accuracy increased for most datasets (five out of six datasets). However, when it was applied to WA-init, the results differed across datasets; accuracy increased in MR and SST-5 datasets, but decreased in CR, TREC, and MPQA datasets. As described in [63], the generated data might be harmful owing to their noise. Indeed, WA-init without data augmentation shows the best performance; its average accuracy on the datasets is 0.7653, whereas those of norm-init, norm-init+LAMBADA, and WA-init+LAMBADA are 0.7448, 0.7528, and 0.7648, respectively. WA-init increases the accuracy of norm-init by a slightly greater margin than LAMBADA. Considering that the data generated may be harmful, WA-init's consistent performance improvements highlight the importance of making good use of existing labeled data.

### D. FURTHER IMPROVEMENT OF WA-INIT

As shown in Figure 2, the WA-init algorithm is trained N times each on two subsets of samples (i.e., $S_A$ and $S_B$). Here, increasing N and the number of subsets $|\mathbb{S}|$ is expected to further improve the performance. In this section, we repeated the experiments of the WA-init algorithm with a higher N and more split subsets. We increased N from 5 to 10. In addition, we randomly split the 2% data four times with a ratio of 5:5 for train/validation sets and obtained four sample subsets, $S_A$, $S_B$, $S_C$, and $S_D$. Thus, Equation (4) for the good initialization can be rewritten as follows: $P_{good} = \frac{(1-\alpha)}{4}(P_A + P_B + P_C + P_D) + \alpha P_{init}$. $\alpha$ was set to the same 1/3. Table 9 presents the results. The accuracy improved for all datasets except the SST-2. Therefore, in WA-init, it is important to train with various combinations of samples and obtain a better model on each sample subset. Section VI-B provides an analysis of why WA-init works well.

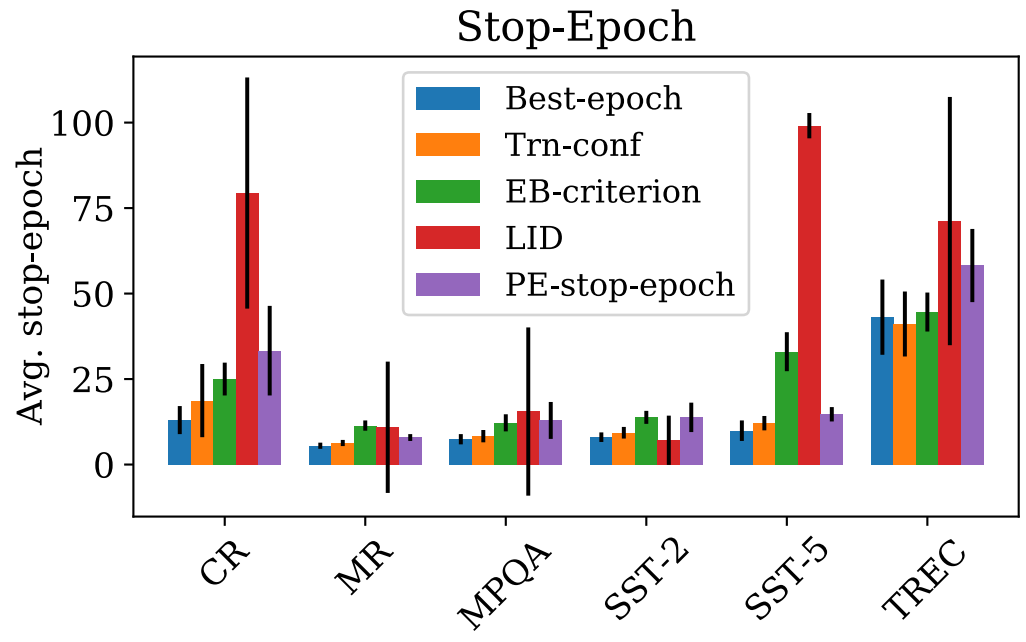

**FIGURE 6. Stopping point comparison in our low-resource setting. The best epoch indicates the point at which the test loss is the lowest during training. The black vertical lines represent a ±1 standard deviation.**

## VI. DISCUSSION

### A. ANALYSIS FOR STOPPING EPOCH

As shown in Table 2, the EB criterion resulted in better accuracy but worse loss and calibration error than the trn-conf algorithm. Figure 5 illustrates why the stop criteria produced these different results. We recorded the test performance at every epoch on the SST-5 dataset. As shown in the left side of Figure 5, the EB criterion met Equation (2) at the latter epoch, whereas the stopping point of the trn-conf was close to the ideal point with the lowest test loss. Additionally, the model tended to be overconfident on the predicted samples as training continued, resulting in a slightly better accuracy but worse loss and calibration error. The confidence histogram (right side of Figure 5) shows that the gap between the test accuracy and average confidence of the trn-conf algorithm was significantly smaller than that of the EB criterion, indicating that the calibration error of the trn-conf algorithm was better.

Figure 6 shows the average stopping points of the non-validation stop criteria. The best epoch indicates the point at which the test loss was the lowest during training. Generally, the stop epoch of the trn-conf algorithm was closer to the best epoch than the other stop criteria. LID showed a significantly different stopping epoch over the datasets and a high standard deviation for each. Thus, LID may not be suitable as a stop method in low-resource NLP tasks. The EB criterion and PE-stop-epoch generally resulted in later stop epochs than the best epoch. The EB criterion uses the gradients of the training samples, which may have difficulty reflecting the evaluation metrics (i.e., accuracy and loss). The PE-stop-epoch is precalculated using the previous validation loss. As such, the PE-stop-epoch algorithm cannot cope with the changes occurring during training, because the epoch is fixed. On the other hand, the trn-conf criterion is calculated during training, and the utilization of the sample confidences is related to the evaluation metric (i.e., loss). The reference confidences of the trn-conf algorithm can thus serve as a stop criterion reflecting the test loss. We speculate that this is why the trn-conf algorithm provided a stopping point close to the best epoch with the lowest test loss.

**TABLE 10.** Initial (*epoch* = 0) performance when the model is initialized with either partially fitted or general parameters. For fair comparison, we ensured that the number of parameters in $P^{gen}$ and $P^{part}$ is the same, because the performance may be affected by the number of parameters.

| Dataset | Param | Acc | Loss | ECE | OE |
|---|---|---|---|---|---|
| SST-2 | $P_{part}$ | 0.608$_{\pm.028}$ | 0.664$_{\pm.027}$ | 0.142$_{\pm.022}$ | 0.062$_{\pm.023}$ |
| | $P_{gen}$ | **0.743**$_{\pm.029}$ | **0.535**$_{\pm.027}$ | **0.106**$_{\pm.010}$ | **0.020**$_{\pm.010}$ |
| Difference | | +0.135 | −0.129 | −0.036 | −0.042 |
| **Dataset** | **Param** | Acc | Loss | ECE | OE |
| SST-5 | $P_{part}$ | 0.267$_{\pm.009}$ | 1.609$_{\pm.011}$ | 0.073$_{\pm.013}$ | 0.022$_{\pm.006}$ |
| | $P_{gen}$ | **0.322**$_{\pm.015}$ | **1.529**$_{\pm.017}$ | **0.043**$_{\pm.006}$ | **0.011**$_{\pm.004}$ |
| Difference | | +0.055 | −0.080 | −0.030 | −0.011 |

**TABLE 11.** Accuracy comparison of WA-init and conf-ensemble algorithms. Conf-ensemble averages the prediction confidences of two models, whose parameters are $P_A$ and $P_B$.

| | CR | MPQA | MR | SST-2 | SST-5 | TREC |
|---|---|---|---|---|---|---|
| Conf-ensemble | 0.800 | 0.847 | 0.794 | 0.839 | 0.387 | 0.807 |
| WA-init | **0.831** | **0.864** | **0.796** | **0.854** | **0.408** | **0.839** |
| Difference | +0.031 | +0.017 | +0.002 | +0.015 | +0.021 | +0.032 |

### B. ANALYSIS OF WA-INIT INITIALIZATION

As shown in Section V-B, the WA-init algorithm resulted in a better average prediction performance than the other initialization methods. In this section, we analyze why the WA-init algorithm worked well. We speculate that the weight averaging over different sample subsets would neutralize partially fitted parameters that may have negative impacts while preserving the general parameters providing positive impacts. Note that the weight averaging applies two parameter sets (i.e., $P_A$ and $P_B$) that are fitted to the sample sets (i.e., $S_A$ and $S_B$), respectively, as shown in Figure 2. After training, each updated parameter is incremented or decremented from the initial parameters. In this weight averaging, we can consider the parameters that increase or decrease simultaneously in both $P_A$ and $P_B$ to be *well generalized*. Conversely, the parameters that are incremented in one parameter set but decremented in the other can be considered *partially fitted* to the sample set: either $S_A$ or $S_B$. In this context, the WA-init algorithm, which averages $P_A$ and $P_B$, preserves the general parameters and neutralizes the partially fitted ones. For example, in partially fitted parameters, the average of the increased parameter in $P_A$ and the decreased parameter in $P_B$ are similar to the original initial parameter in $P_{init}$. Thus, the WA-init algorithm can improve performance.

One experimental result in which we compared the initial accuracies of general and partially fitted parameters supports this speculation. The initial accuracy refers to the accuracy at the zero epoch after being initialized with a parameter set. We distinguished the general and partially fitted parameters in $P_A$, denoted as $P^A_{gen}$ and $P^A_{part}$, respectively. Then, we initialized one model with $P^A_{gen}$ and the other with $P^A_{part}$. We conducted the same procedure for $P_B$ and then averaged the results of $P^A_{gen}$ and $P^B_{gen}$ and those of $P^A_{part}$ and $P^B_{part}$. Table 10 shows the results on the SST-2 and SST-5 datasets. As shown in Table 10, the general parameters outperformed the partially fitted ones by a large margin. It can be seen that the general parameters are better generalized to the dataset than the partially fitted ones.

Second, we speculate that in the WA-init algorithm, the model would learn new knowledge between the $S_A$ and $S_B$ while training all the samples, $S = S_A \cup S_B$. We fed test data into two models, whose respective parameters were $P_A$ and $P_B$, and averaged the two prediction confidences. Note that this algorithm is a simple confidence ensemble (conf-ensemble) of two independent models, not an initialization method. Although the WA-init and conf-ensemble algorithms are similar, Table 11 shows that the WA-init algorithm outperforms the conf-ensemble method for all datasets. This result supports our speculation and highlights the importance of the training strategy.

### C. EFFICACY OF THE VALIDATION SET IN LOW-RESOURCE SETTINGS

Validation sets are used to measure prediction performance, based upon which the validation set can be used for early stopping, hyperparameter tuning, and searching for initialization parameters. As shown in Section V-A, early stopping based on a separate validation set did not show good prediction performance owing to the reduced training samples. However, it also showed that the validation set can be exploited to obtain other stopping criteria, training confidences and the averaged stop epoch, which were used in the trn-conf and PE-stop-epoch algorithms, respectively. Thereafter, training all the labeled samples resulted in better prediction performance. In Section V-B, the validation set was useful in finding better initial parameters. The greedy-init and WA-init algorithms searched for good parameters based on validation performance. Meanwhile, Figure 7 presents the performance of the val-N-runs method according to the N in the CR dataset. The performance consistently increased with the number of runs. There was a very large performance difference between the val-75 and -25%-N-runs methods, indicating that in low-resource settings, it is generally better to use as much training data as possible with a small validation set. When N was greater than 2, the performance of the val-75%-N-runs method was better than training all samples with the norm-init initialization. However, the WA-init algorithm showed better performance than the val-75%-N-runs method, even when N was 10. Note that the training times of the WA-init algorithm and val-75%-10-runs method are similar in our settings. These results suggest the importance of effectively using the validation set.

The validation set can also be used for hyperparameter tuning. However, a small validation set may not provide well-tuned hyperparameters. Figure 8 compares the two results of one model trained with default hyperparameters and another with tuned hyperparameters. The default hyperparameter setting was the same as that used in this paper, in which the learning rate was $2e^{-5}$, drop-out was 0.1, and batch size was 32. For hyperparameter tuning, we used half of the labeled data for validation. After tuning, all samples were trained using the tuned hyperparameters. We used the norm-init distribution for initialization and the trn-conf algorithm for early stopping. We greedily tuned within the range [$1e^{-5}$, $2e^{-5}$, $4e^{-5}$, $8e^{-5}$] for the learning rate, [0.1, 0.3] for the drop-out, and [16], [32] for the batch size. As shown in Figure 8, the tuned hyperparameters were slightly better (1.1% improvement) in only the MPQA dataset, whereas worse performance was shown by the remaining five. The default setting showed at least 0.7% and at most 2.8% higher accuracy than the tuned hyperparameters. This experiment demonstrates that the validation set may not be adequate for the hyperparameter tuning in low-resource settings.

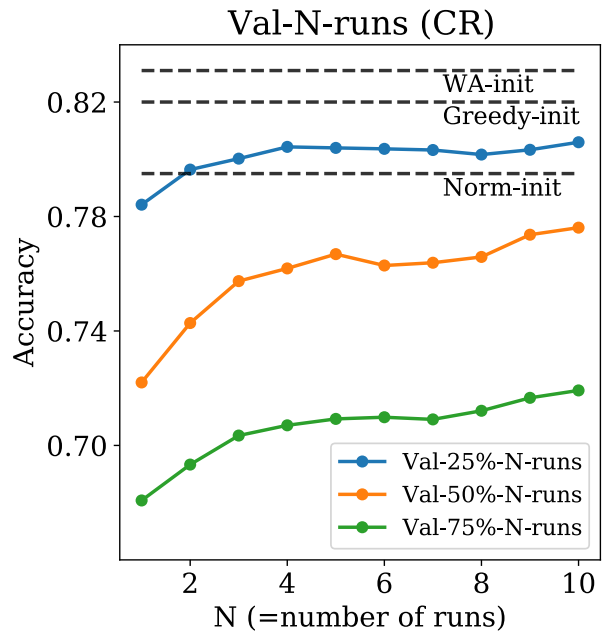

**FIGURE 7.** Accuracy of the val-N-runs method according to N, the number of runs. Each model stops at the lowest validation loss during N runs.

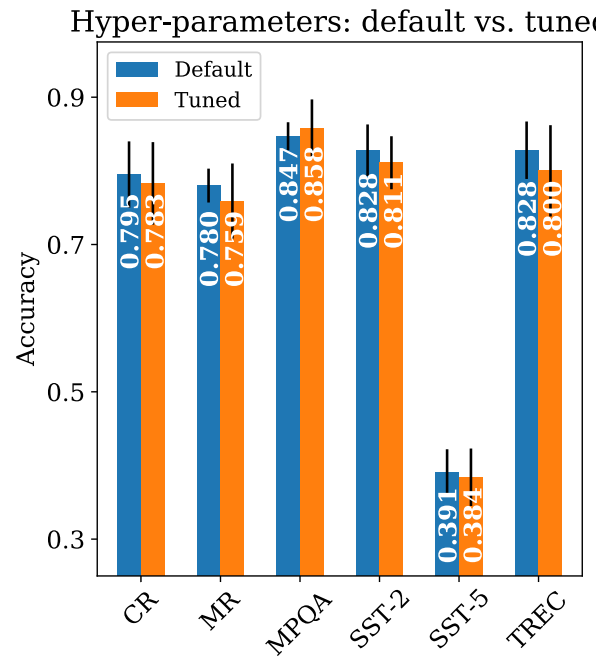

**FIGURE 8.** Hyperparameters comparison of default and tuned settings in low-resource datasets.

**TABLE 12.** Memory and time complexities when training. $b$ is the batch size and $s$ is the sentence length; these were set to 32 and 64, respectively. As shown in Figure 3, $d_{ff}$ is the dimension of the feed-forward layer, and $n_h$ and $n_l$ denote the number of heads and layers, respectively. *The complexity of BERT follows that of the transformer model [54], [64]. The running time and GPU memory were measured per batch. Our computational environment is presented in Section IV-C. Note that our approach, WA-init, does not increase the complexities when testing.

| Model | Memory Complexity | Time Complexity |
|---|---|---|
| BERT* | $\max(bsd_{ff}, bn_h s^2)n_l$ | $(bsd_{ff}+bn_h s^2)n_l$ |
| *Measured results in our environment.* | | |
| BERT-base (baseline) | 4,084 (MB) | 0.285 (s) |
| WA-init (N=5,$\lvert\mathbb{S}\rvert$=2) | 4,084 (MB) | 3.132 (s) |

### D. LIMITATIONS OF OUR STUDY

This study explored various techniques that can be applied in the paucity of labeled data. Among these techniques, the WA-init with trn-conf stop method showed significantly higher performance compared to other methods. However, our explored methods, including the WA-init and trn-conf, have the following limitations. First, they need to tune all the parameters of the network. Such full fine-tuning requires high memory capacity; it can be inefficient for larger models. Second, in the case that a learning model cannot appropriately represent the prediction probability (i.e., confidence) or its training and testing are calculated in a different way, the stop criterion trn-conf may not work well. For example, the PERFECT model presented in Section V-C has different processes when training and testing [58]. Third, the WA-init method repeats training multiple times, resulting in a longer running time. However, taking several more times of the training time might not be significantly burdensome in low resource settings, because the amount of data to train is small. Table 12 shows the memory and time complexities. In our settings, WA-init requires approximately 11 times of more training time than the baseline; however, it is a sufficiently viable time. It is also noted that our approach does not increase GPU usage and test time.

Several techniques may mitigate these limitations. For example, for the first limitation, applying several layers that allow tuning only few additional parameters may reduce the memory usage [65]. For the second limitation, researchers can refer to the results in Section V-A and select other suitable stop criteria such as EB criterion and PE-stop-epoch. In the future, we hope that several methods will be studied to deal with these limitations.

## VII. CONCLUSION

There have been many low-resource algorithms that utilized additional resources or required model manipulation. However, few studies have examined methods to improve performance based on only a small amount of labeled data, without model redesign. In this study, we explored several methods that are easily implementable with only a few labeled data and without model manipulation. To this end, we focused on training-validation splitting, early stopping, and initialization methods, and we experimented on six sentence classification datasets. The experimental results imply that a large validation set in a low-resource setting can be detrimental. For early stopping, the performances significantly differed by the stop criteria and evaluation metrics. For initialization methods, the WA-init algorithm showed the best performance in most datasets. The WA-init algorithm was also effectively adapted to various SOTA models, resulting in further performance improvement. These results demonstrate that the WA-init algorithm, a simple integrated method, is effective in limited-data regimes. Our results highlight the importance of the training strategy in low-resource settings. In the future, we expect that our approach will be applied to real low-resource data such as African languages or other fields, such as computer vision. We also hope that our approach will improve in terms of parameter efficiency. To this end, we plan to study how to perform good initialization using only few parameters. Future studies are expected to enable more effective utilization of large-scale pre-trained models by updating fewer parameters, thereby increasing their applicability across various domains.

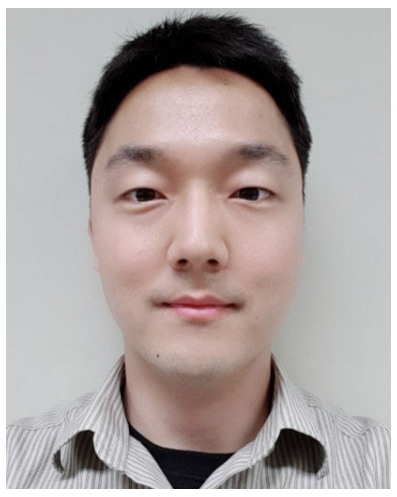

**HONGSEOK CHOI** received the B.S. degree in information and communication engineering from Inha University, Incheon, South Korea, in 2016. He is currently pursuing the Ph.D. degree in electrical engineering and computer science with the Gwangju Institute of Science and Technology, Gwangju, South Korea.

He was discharged as a Sergeant in the Army, in 2013. His research interests include machine learning, deep learning, and natural language processing.

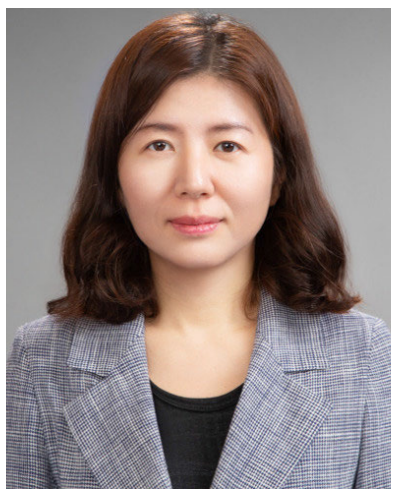

**HYUNJU LEE** (Member, IEEE) received the B.S. degree in computer science from the Korea Institute of Science and Technology, South Korea, in 1997, the M.S. degree in computer engineering from Seoul National University, South Korea, in 1999, and the Ph.D. degree in computer science from the University of Southern California, Los Angeles, CA, USA, in 2006.

From 2006 to 2007, she was a Postdoctoral Research Fellow with Brigham Women's Hospital, Harvard Medical School. Since 2007, she has been with the School of Electrical Engineering and Computer Science and the Artificial Intelligence Graduate School, Gwangju Institute of Science and Technology, Gwangju, South Korea. Her research interests include machine learning, natural language processing, and bioinformatics.